\documentclass[12pt, a4paper, twoside]{article}
\usepackage[utf8]{inputenc}
\usepackage[textwidth=15cm]{geometry}
\usepackage{graphicx}
\usepackage{subfigure}
\usepackage{amsmath}
\usepackage{comment,authblk,natbib,graphicx}
\usepackage{caption}
\usepackage{ragged2e}
\usepackage{algorithm}
\usepackage{algpseudocode}
\usepackage{array}
\usepackage{appendix}
\usepackage{amssymb}
\usepackage{makecell}
\title{Regularized Soft Actor-Critic for Behavior Transfer Learning\footnote{paper accepted by IEEE CoG2022}}

\author[1]{Mingxi Tan\thanks{ming-xi.tan@ubisoft.com}}
\author[2]{Andong Tian\thanks{an-dong.tian@ubisoft.com}}
\author[3]{Ludovic Denoyer\thanks{ludovic.denoyer@ubisoft.com}}

\affil[1,2,3]{La Forge, Ubisoft}

\date{}

\begin{document}

\maketitle

\begin{abstract}
Existing imitation learning methods mainly focus on making an agent effectively mimic a demonstrated behavior, but do not address the potential contradiction between the behavior style and the objective of a task. There is a general lack of efficient methods that allow an agent to partially imitate a demonstrated behavior to varying degrees, while completing the main objective of a task. In this paper we propose a method called Regularized Soft Actor-Critic which formulates the main task and the imitation task under the Constrained Markov Decision Process framework (CMDP). The main task is defined as the maximum entropy objective used in Soft Actor-Critic (SAC) and the imitation task is defined as a constraint. We evaluate our method on continuous control tasks relevant to video games applications.
\end{abstract}

\section{Introduction}
In video games, non-playable characters (NPCs) and bots are usually expected to complete a mission while adopting specific behavior styles. Traditional ways require to build complex behavior trees, which can be a challenging and time-consuming work. We therefore propose to treat this problem as a behavior imitation learning problem for Reinforcement Learning. In recent years, many behavior imitation learning algorithms have been proposed \citep{Pomerleau, Ng_Russell,Ho_Ermon, Fu_Luo_levine, Reddy_Dragan_levine, Ross_Gordon_Bagnell}. These algorithms, such as behavioral cloning \citep{Ross_Gordon_Bagnell}, inverse reinforcement learning \citep{Ho_Ermon}, general adversarial imitation learning \citep{Fu_Luo_levine} and soft Q imitation learning \citep{Reddy_Dragan_levine}, mainly focus on how to make an agent mimic the entire demonstrated behaviors rather than partially imitating the demonstrated behaviors to varying degrees. However, it is a common need in video games to have NPCs behaving in a specific style while still accomplishing their main tasks.

Although this need can be addressed with a multi-objective framework, where the first objective is to accomplish the main task and the second objective is to imitate the demonstrated behavior. An effective multi-objective framework requires explicit constant rewards for both the main task and the imitation task, which can be time-consuming to design as it usually requires a reward-shaping loop to find suitable values \citep{Mossalam_Assael_Roijers,Abels_Roijers_Lenaerts, yang_sun}.

In this work we focus on making the agent partially imitate the demonstrated behavior to varying degrees while completing a main task by formulating the imitation learning under the Constrained Markov Decision Processes (CMDPs) framework. We propose to extend the Soft Actor-Critic (SAC) \citep{Haarnoja_Tang} to an efficient partial imitation learning algorithm named Regularized Soft Actor-Critic (RSAC). We evaluate our algorithm on a Rabbids\footnote{https://en.wikipedia.org/wiki/Raving\_Rabbids} video game, which provides a relevant context to the application of reinforcement learning to the design of NPCs behavior styles.

\section{\textbf{Background}}
\subsection{\textbf{Markov Decision Process (MDPs)}}
The Markov decision processes (MDPs) defined by the tuple \((S; A; p_s; r)\) where \(S\) denotes the state space, \(A\) the action space, \(p_s(s'|s, a)\) the transition distribution, \(r(s_t,a_t)\) the reward function. The agent's policy \(\pi\) is a state-conditional distribution over actions, where \(\pi(a_{t}|s_{t})\) denotes the probability of taking action \(a_{t}\) in state \(s_{t}\). We will use \(\rho_\pi(s_t)\) and \(\rho_\pi(s_t,a_t)\) to denote the state and state-action marginals of the trajectory distribution induced by \(\pi(a_t|s_t)\). Giving any scalar function of actions and states  \(f:S\times A \to \mathbb{R}\), the expected discounted sum of f is defined as
\begin{equation} \label{eq1}
\begin{aligned}
& J(\pi):= E_{(s_t,a_t)\sim\rho_\pi}\left[\sum\limits_{t=0}^{T}\gamma^t f(s_t, a_t)\right],
\end{aligned}
\end{equation}
where \(\gamma \in [0,1]\) is a discount factor used to reduce the relative importance of future rewards. In Reinforcement learning, the goal is to learn a policy \(\pi(a_t|s_t)\) that maximizes the reward function over the whole trajectory 
\begin{equation} \label{eq2}
\begin{aligned}
\pi^*= \mathop{argmax}\limits_{\pi\in\Pi}\left[E_{(s_t,a_t)\sim\rho_\pi}\sum\limits_{t=0}^{T}r(s_t, a_t)\right],
\end{aligned}
\end{equation}
where \(\Pi\) is the set of possible policies.

\subsection{\textbf{Constrained Markov Decision Process (CMPDs)}}
Constrained Markov decision processes (CMDPs) restrict the policies set \(\Pi\) to a typically smaller set \(\Pi_{C}\) by introducing the set of constraints constructed by cost functions \(C_k:S\times A \to \mathbb{R}\) and their corresponding thresholds \(d_k\in\mathbb{R}\), with \(k = 1, ..., K\). The goal of Reinforcement Learning under CMDPs is to find a plausible policy by solving the constrained optimization problem
\begin{equation} \label{eq3}
\begin{split}
&\pi^*= \mathop{argmax}\limits_{\pi\in\Pi_{C}}\left[E_{(s_t,a_t)\sim\rho_\pi}\sum\limits_{t=0}^{T}r(s_t, a_t)\right]\\
&s.t.\quad  J_{C_{k}}(\pi) \leq d_{k}, k=1,..K,\\
\end{split}
\end{equation}
where \(\Pi_{C}\) is the set of possible policies that satisfy all constraints.

\section{\textbf{Our Approach}}
\subsection{\textbf{Problem Setting}}
In this paper, we focus on finding the policy that allows the agent to complete the task while having a similar behavior as the behavior shown in a demonstration clip. The similarity between the behavior of the agent and the behavior in the demo clip can be defined by the cross-entropy between the policy of the agent \(\pi\) and the policy of the demo clip \(\pi^{d}\). \(\pi^{d}\) is trained by following the imitation learning algorithm of \citep{Reddy_Dragan_levine}. Then our problem can be formulated as a policy search in CMDPs. The objective can be defined as a maximum entropy objective following the prior work of \citep{Haarnoja_Tang} and the constraint can be defined by the cross-entropy between the policy of the agent and the policy of the behavior in the demo clip
\begin{equation} \label{eq4}
\begin{split}
&\mathop{max}\limits_{\pi_{0:T}}E_{(s_t,a_t)\sim\rho_\pi}\left[\sum\limits_{t=0}^{T}\left[r(s_t, a_t) + \alpha H(\pi(\cdot|s_t))\right]\right]\\
&s.t.\quad  E_{(s_t,a_t)\sim\rho_\pi}\left[\log\pi^{d}(a_t|s_t)\right]\ge\overline{CE},\\
\end{split}
\end{equation}
where \(H(\pi(\cdot|s_t))\) is entropy of \(\pi\), \(\alpha\) is temperature parameter of entropy term, \(\overline{CE}\) is the required cross-entropy between the policy of the agent and the policy of the demo clip.
For every time step, the \eqref{eq4} can be presented as a dual problem 
\begin{equation} \label{eq5}
\begin{split}
&\mathop{max}\limits_{\pi_{0:T}}E_{(s_t,a_t)\sim\rho_\pi}\left[r(s_t, a_t) + \alpha H(\pi(\cdot|s_t))\right] = \\
&\mathop{min}\limits_{\beta_t\ge 0}\mathop{max}\limits_{\pi_{0:T}}E_{(s_t,a_t)\sim\rho_\pi} \left[\vphantom{\overline{CE}} r(s_t, a_t) + \alpha H\left(\pi(\cdot|s_t)\right) - \beta CE\left(\pi(\cdot|s_t), \pi^{d}(\cdot|s_t)\right) - \beta \overline{CE} \right],\\
\end{split}
\end{equation}
where \(\beta\) is a Lagrangien multiplier. 

We use the strong duality, which holds because the objective function is convex, and the constraint (Cross-Entropy) is also a convex function in \(\pi_t\). This dual objective can be considered as the extended version of the maximum entropy objective of \citep{Haarnoja_Tang}, denoted as regularized maximum entropy objective (RMEO) with respect to the policy as
\begin{equation} \label{eq6}
\begin{aligned}
&\pi^*_t=\mathop{argmax}\limits_{\pi\in\Pi}E_{(s_t,a_t)\sim\rho_\pi}\left[r(s_t, a_t) + \alpha H\left(\pi(\cdot|s_t)\right) - \beta_t CE\left(\pi(\cdot|s_t), \pi^{d}(\cdot|s_t)\right)\right].\\
\end{aligned}
\end{equation}
Note the optimal policy at time t is a function of the dual variable \(\beta_t\). For every time step, we can firstly solve the optimal policy \(\pi_t\) at time t and then solve the optimal variable \(\beta_t\) as
\begin{equation} \label{eq7}
\beta^*_t = \mathop{argmax}\limits_{\beta_t\ge 0}E_{(s_t,a_t)\sim\pi^*_t}\left[\beta_t log\pi^{d}_t(a_t|s_t)\right]-\beta_t \overline{CE}.\\
\end{equation}
To optimize the RMEO in \eqref{eq6}, we first derive a tabular Q-iteration method (RMEO-Q), then present RMEO actor-critic (RMEO-AC), a practical deep reinforcement learning algorithm.

\subsection{\textbf{Regularized Maximum Entropy Reinforcement Learning}}
Following on a similar logic as \citep{Haarnoja_Tang}, the regularized soft Q-function and regularized value function can be defined as
\begin{footnotesize}
\begin{equation} \label{eq8}
\begin{aligned}
&Q^*_{rs}(s_t,a_t) = r_t + E_{(s_{t+1},...)\sim\rho_\pi}\left[\sum\limits_{l=1}^{\infty}\gamma^l\left(r_{t+l}+ \alpha H\left(\pi^*(\cdot|s_{t+l})\right) -\beta CE\left(\pi(\cdot|s_{t+l}), \pi^{d}(\cdot|s_{t+l})\right) \right) \vphantom{\sum\limits_{l=1}^{\infty}} \right].\\
\end{aligned}
\end{equation}
\end{footnotesize}
\begin{equation} \label{eq9}
{
\begin{aligned}
& V^*_{rs}(s_t) = \alpha log\sum\limits_{a'_t}\left(\pi^{d}(a'_t|s_t)^{\frac{\beta}{\alpha}}\exp{\left(\frac{1}{\alpha} Q^\pi_{rs}(s_t, a'_t)\right)} \right).\\
\end{aligned}
}
\end{equation}
The optimal policy of \eqref{eq6} is given by 
\begin{equation} \label{eq10}
\begin{aligned}
\pi^* & =\frac{\left(\pi^{d}(a'_t|s_t)\right)^{\frac{\beta}{\alpha}}\exp{\left(\frac{1}{\alpha}Q^\pi_{rs}(s_t, a'_t)\right)}}{\sum\limits_{a'_t}\left(\pi^{d}(a'_t|s_t)^{\frac{\beta}{\alpha}}\exp{\left(\frac{1}{\alpha} Q^\pi_{rs}(s_t, a'_t)\right)} \right)}\\
& =\left(\pi^{d}(a'_t|s_t)\right)^{\frac{\beta}{\alpha}}exp{\left(\frac{1}{\alpha}Q^\pi_{rs}(s_t, a'_t)-\frac{1}{\alpha}V_{rs}^*(s_t)\right)}.\\
\end{aligned}
\end{equation} 
Proof. See Appendix A.2\\
Then we define the regularized soft Bellman equation for the regularized soft state-action value function Q as\\
\begin{equation} \label{eq11}
Q^*_{rs}(s_t,a_t)=r_t+\gamma E_{(s_{t+1})\sim p_s}\left[V_{rs}(s_{t+1})\right].\\
\end{equation}
Proof. See Appendix A.3\\
Let \(Q^*_rs(\cdot,\cdot)\) and \(V^*_{rs}(\cdot)\) be bounded and assume that \(\sum\limits_{a'_t}\left(\pi^{d}(a'_t|s_t)^{\frac{\beta}{\alpha}}\exp{\left(\frac{1}{\alpha} Q^\pi_{rs}(s_t, a'_t)\right)}\right)\leq\infty\) exists, we can find a solution to \eqref{eq10} with a fixed-point iteration, which we call regularized soft Q-iteration as
\begin{equation} \label{eq12}
{
\begin{aligned}
&Q_{rs}(s_t,a_t) \leftarrow r_t+\gamma E_{(s_{t+1)}\sim p_s}\left[V_{rs}(s_{t+1})\right], \forall s_t, a_t,\\
&V_{rs}(s_t) \leftarrow \alpha log\sum\limits_{a'_t}\left( \vphantom{\exp{\left(\frac{1}{\alpha} Q^\pi_{rs}(s_t, a'_t)\right)}} \pi^{d}(a'_t|s_t)^{\frac{\beta}{\alpha}} \exp{\left(\frac{1}{\alpha} Q^\pi_{rs}(s_t, a'_t)\right)} \right), \forall s_t,\\
\end{aligned}
}
\end{equation}
converge to the optimal \(Q^*_{rs}\) and \(V^*_{rs}\).
Proof. See Appendix B.3\\

To make RMEO-Q effectively imitate the behavior of the demo clip, we use two replay buffers \(D^{d}\) and \(D^{new}\) to store the demonstration experiences and new experiences and use them to update \(Q_{rs}(s_t,a_t)\) separately. We firstly use the new experiences to update \(Q_{rs}\), allowing the agent to learn to complete the task by setting \(\beta\) to zero. Secondly, we continually use the same new experiences to solve \(\beta\) by \eqref{eq7}. If the cross-entropy of \(\pi\) and \(\pi^{d}\) is bigger than the required value (\(\overline{CE}\)), \(\beta\) will be augmented. Otherwise, \(\beta\) will be gradually decreased to zero. Then we use this \(\beta\) and the demonstration experiences to update \(Q_{rs}\), allowing the agent to learn to imitate the demonstrated behavior. 
A large \(\beta\) encourages the agent to learn to mimic the demonstrated behavior, and a small \(\beta\) encourages the agent to learn to complete the task. Thus \(\beta\) works as an automatic regulator that can help find the policy to satisfy both the task completion and the behavior imitation.

\subsection{\textbf{Regularized Maximum Entropy Objective Actor-Critic}}
To practically implement our method, we propose the RMEO actor-critic (RMEO-AC), which uses neural networks as function approximators for both the regularized Q-function and policy and optimizes both networks with stochastic gradient descent. 
We parameterize the regularized Q-function and policy by \(Q_{\theta}(s_t,a_t)\) and \(\pi_{\phi}(a_t|s_t)\).

Then we can replace the Q-iteration with Q-learning and train \(\theta\) to minimize 
\begin{equation} \label{eq13}
{
\begin{aligned}
&J_Q(\theta)=E_{(s_t,a_t)\sim D}\left[\frac{1}{2}\left(Q_{\theta}(s_t,a_t)-r(s_t,a_t) - \gamma E_{(s_{t+1})\sim p_s}[\overline{V}_{rs}(s_{t+1})]\right)^2\right]\\
&V_{rs}(s_{t+1})=Q_{\overline{\theta}}(s_{t+1},a_{t+1})-\alpha log\left(\pi_{\phi}(a_{t+1}|s_{t+1})\right)+\beta log\left(\pi^{d}(a_{t+1}|s_{t+1})\right),\\
\end{aligned}
}
\end{equation}
\\
where
\(Q_{\overline{\theta}}\) is target function for providing relatively stable target values. 
We can also find \(\beta\) by minimizing
\begin{equation} \label{eq14}
{
J(\beta)=E_{(s_t,a_t)\sim D^{new}}\left[\beta \pi^{d}(a_t|s_t)-\beta \overline{CE}\right],\\
}
\end{equation}
The policy parameters can be learned by directly minimizing the KL-divergence
\begin{equation} \label{eq15}
{
\begin{aligned}
&J_\pi(\phi)=E_{(s_t)\sim D}\left[\vphantom{\frac{\left(\pi^{d}(\cdot|s_t)\right)^{\frac{\beta_t}{\alpha}}\exp{\left(\frac{1}{\alpha}Q^\pi_{rs}(s_t,\cdot)\right)}}{Z_{\theta}(s_t)}} D_{KL}\left(\vphantom{\frac{\left(\pi^{d}(\cdot|s_t)\right)^{\frac{\beta}{\alpha}}\exp{\left(\frac{1}{\alpha}Q^\pi_{rs}(s_t,\cdot)\right)}}{Z_{\theta}(s_t)}} \pi_{\phi}(\cdot|s_t)\left|\right|\frac{\left(\pi^{d}(\cdot|s_t)\right)^{\frac{\beta_t}{\alpha}}\exp{\left(\frac{1}{\alpha}Q^\pi_{rs}(s_t,\cdot)\right)}}{Z_{\theta}(s_t)}\right)\right],\\
\end{aligned}
}
\end{equation}
\\
Since the partition function \( Z_\theta\left(s_t\right)\) normalizes the distribution, it does not contribute to the gradient and can be ignored. 
By using the reparameterization trick \(a_t= f_{\phi}(\epsilon;s_t)\), we can rewrite the objective above as
\begin{equation} \label{eq16}
\begin{aligned}
&J_{\pi}(\phi)=E_{(st)\sim D,\epsilon \sim N}\left[\vphantom{\beta log\left(\pi^{d}(f_{\phi}(\epsilon_t;s_t)|s_t)\right)}\alpha log\left(f_{\phi}(\epsilon_t;s_t)|s_t\right)-\beta log\left(\pi^{d}(f_{\phi}(\epsilon_t;s_t)|s_t)\right)-Q_{\theta}\left(s_t,f_{\phi}(\epsilon_t;s_t)\right)\right]\\
\end{aligned}
\vspace{1.0em}
\end{equation}

The gradient of \eqref{eq13}, \eqref{eq14}, \eqref{eq15} with respect to the corresponding parameters are
\begin{equation} \label{eq17}
{\nabla_{\theta}J_Q(\theta)=\nabla_{\theta}Q_{\theta}(s_t,a_t)\left(Q_{\theta}(s_t,a_t)- r(s_t,a_t)-\gamma \overline{V}(s_{t+1})\right)
},
\end{equation}
\vspace{1em}
\begin{equation} \label{eq18}
{
\nabla_{\beta}J(\beta)=log\pi^{d}(a_t|s_t)-\overline{CE},
}
\end{equation}
\begin{equation} \label{eq19}
\resizebox{0.99 \textwidth}{!}{%
$
\nabla_{\phi}J_{\pi}(\phi)=\nabla_{\phi}\alpha log\left(\pi_{\phi}(a_t|s_t)\right)+ \nabla_{\phi}f_{\phi}(\epsilon_t;s_t)\left(\nabla_{a_t}\alpha log(\pi_{\phi}(a_t|s_t))-\nabla_{a_t}\beta log(\pi^{d}(a_t|s_t))-\nabla_{a_t}Q_{\theta}(s_t,a_t)\right),\\
$
}
\vspace{1em}
\end{equation}

The final algorithm is listed in Algorithm 1.\\
\begin{scriptsize}
{\noindent} \rule[0.25\baselineskip]{\textwidth}{1.5pt}\\
\textbf{Algorithm 1} Regularized Soft Actor-Critic\\
{\noindent} \rule[0.25\baselineskip]{\textwidth}{0.5pt}\\
\textbf{Input:} \(\theta, \phi, \beta, \pi^{d}, D^{d}\) \hfill Initial parameters, policy and replay buffer of demonstrated behaviors\\
\hspace*{1cm}\(\overline{\theta}\leftarrow \theta\) \hfill Initialize target network weights\\
\hspace*{1cm}\(D_{new}\leftarrow \emptyset\) \hfill Initialize an empty replay buffer for new experiences\\
\hspace*{1cm}\textbf{for} each iteration \textbf{do}\\
\hspace*{2cm}\textbf{for} each environment step \textbf{do}\\
\hspace*{3cm}\(a_t\sim \pi_{\phi}(a_t|s_t)\) \hfill Sample action from the policy\\
\hspace*{3cm}\(s_{t+1}\sim P(s_{t+1}|s_t,a_t)\) \hfill Sample transition from the environment\\
\hspace*{3cm}\(D^{new} \cup \left\{(s_t,a_t,r(s_t,a_t), s_{t+1})\right\}\) \hfill store the new transition in replay pool\\
\hspace*{2cm}\textbf{end for}\\
\hspace*{2cm}\textbf{for} each gradient step \textbf{do}\\
\hspace*{3cm}Sample a mini-batch from \(D^{new}\)\\
\hspace*{3cm}first set \(\beta = 0\)\\
\hspace*{3cm}\(\theta\leftarrow\theta-\lambda_Q \nabla_{\theta}J_Q(\theta)\) \hfill update the Q-function parameters\\
\hspace*{3cm}\(\phi\leftarrow\phi-\lambda_{\pi}\nabla_{\phi}J_{\pi}(\phi)\) \hfill update policy weights\\
\hspace*{3cm}\(\overline{\theta}\leftarrow\tau\theta+(1-\tau)\overline{\theta}\) \hfill update target network weights\\
\hspace*{3cm}then \(\beta\leftarrow\beta-\nabla_{\beta}J(\beta)\) \hfill update \(\beta\)\\
\hspace*{3cm}Sample a mini-batch from \(D^{d}\)\\
\hspace*{3cm}\(\theta\leftarrow\theta-\lambda_Q\nabla_{\theta}J_Q(\theta)\) \hfill update the Q-function parameters\\
\hspace*{3cm}\(\phi\leftarrow\phi-\lambda_{\pi}\nabla_{\phi}J_{\pi}(\phi)\) \hfill update policy weights\\
\hspace*{3cm}\(\overline{\theta}\leftarrow\tau\theta+(1-\tau)\overline{\theta}\) \hfill update target network weights\\
\hspace*{2cm}\textbf{end for}\\
\hspace*{1cm}\textbf{end for}\\
Output: \(\theta, \phi\)\\
{\noindent} \rule[0.25\baselineskip]{\textwidth}{1.5pt}
\end{scriptsize}

\section{\textbf{Experiments}}
\begin{figure}[b]
\centering
\subfigure[Main task]{\includegraphics[width=7.4cm]{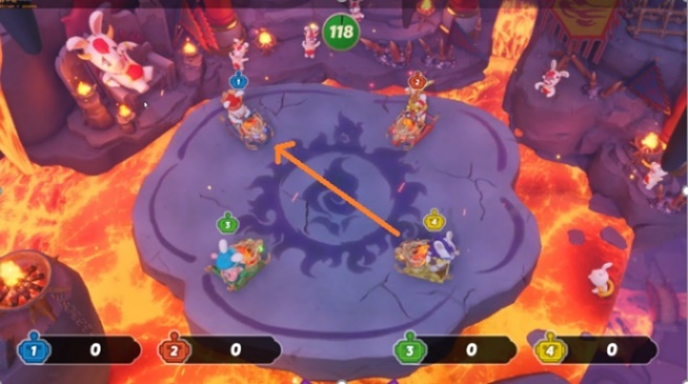}}
\subfigure[Behavior of demo clip]{\includegraphics[width=7.4cm]{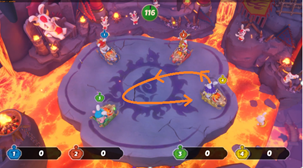}}
\caption{Main task and behavior of demo clip}
\label{fig1}
\end{figure}

To evaluate the proposed approach, we first define the main task and then define the specific behavior that should be imitated. The agent we train is car number 4 at the bottom-right corner of the map and its main task is to hit car number 1 at the top-left corner as shown in Fig.~\ref{fig1} (left). The demonstrated behavior is car number 4 navigating in circles and backwards in the lower half of the map as shown in Fig.~\ref{fig1} (right). The goal of the agent is to hit car number 1 while adopting the demonstrated behavior style. This task is challenging because if the agent mimics the behavior of the entire demonstration, it cannot hit car number 1. It requires car 4 to partially mimic the behavior shown while hitting car 1. 

\subsection{\textbf{Baseline}}
We modify the Soft Q Imitation Learning algorithm of \citep{Reddy_Dragan_levine} by adding a constant main task reward +1.0. We perform a grid search on 11 imitation reward values: 0.0, +0.1, +0.2, +0.3, +0.5, +0.6, +0.7, +0.8, +0.9,+1.0 in the environment of a Rabbids video game. Since the reward for the main task is sparse, we use an auxiliary reward to guide the agent to complete the main task: when the agent is in the upper half of the map, if it is far from car 1, it will obtain a constant reward = -0.01. For each reward setting, we train the model for 1M steps using the same random seed, and then evaluate the model for 10 epochs. The cross-entropy and reward for each setting are shown in Table \ref{tab1}. The first row is different imitation reward applied in experiment, the second row is the corresponding cross-entropy of \(\pi\) and \(\pi^{d}\) and the third row is the corresponding task rewards obtained. All values are the average over 10 episodes.

\begin{center}
\begin{table}[t]
        \caption{Results of baseline models.}
        \centering
        \normalsize
        \resizebox{.9 \textwidth}{!}
        {
        \begin{tabular}{|c|c|c|c|c|c|c|c|c|c|c|c|}
            \hline
            \makecell{Imitation\\ reward} & 0.0 & 0.1 & 0.2 & 0.3 & 0.4 & 0.5 & 0.6 & 0.7 & 0.8 & 0.9 & 1.0 \\
            \hline
            CE & 1.72 & 1.48 & 1.44 & 1.44 & 1.44 & 1.44 & 1.44 & 1.43 & 1.43 & 1.43 & 1.43 \\ 
            \hline
            \makecell{Mission\\reward} & 0.96 & 0.92 & -0.24 & -0.26 & -0.31 & -0.35 & -0.38 & -0.42 & -0.41 & -0.42 & -0.43\\ 
            \hline
        \end{tabular}
        }
        \label{tab1}
    \label{tab:t4}
\end{table}
\end{center}
From the Table \ref{tab1}, we make several important observations: 1) Except for the model trained with imitation reward=0.0 and reward=+0.1, the cross-entropy of all other models is in a very narrow range (second row), which means it is difficult to tune the constant reward of the imitation task to obtain an agent which behavior has a certain degree of similarity to the demonstrated behavior. 2) Out of 11 imitation reward settings, only one setting (imitation reward=+0.1) successfully trained a policy that can complete the main task in the demonstrated behavior style, which means the reward shaping is very challenging and time-consuming.

\subsection{\textbf{Experiments settings}}
\begin{figure}[htbp]
\centering
\subfigure[ CE of \(\pi\) and \(\pi^{d}\)]{\includegraphics[width=4.7cm]{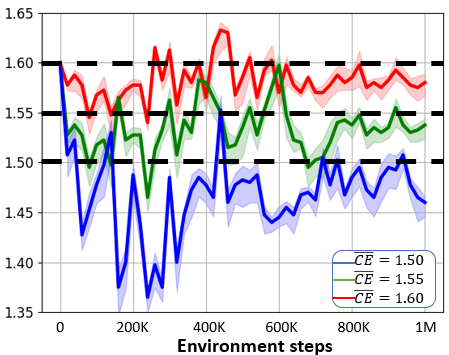}}
\subfigure[Total reward of three agents]{\includegraphics[width=4.7cm]{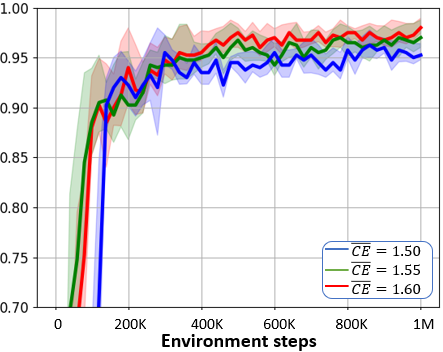}}
\subfigure[\(\beta\) of three agents]{\includegraphics[width=4.7cm]{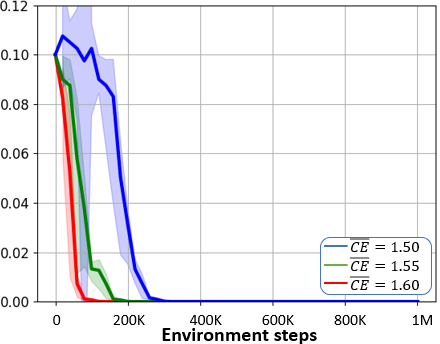}}
\caption{Results of our approach under three different constraints (\(\overline{CE}\)). a) CE of \(\pi\) and \(\pi^{d}\). b) Total reward of three agents. c) Three \(\beta\) values. All curves show the average over 5 random seeds and envelops show the standard error around the mean.}
\label{fig2}
\end{figure}

We train three agents to complete the main task while partially imitating the demonstrated behavior to 3 different degrees: cross-entropy of \(\pi\) and \(\pi^{d}\) smaller than 1.5, 1.55 and 1.6. When the agent completes the main task, it will get a constant reward = +1.0. We apply the same auxiliary reward and train the same steps as before. For every setting, we train the model with 5 random seeds, we halt the training process every 20K steps and evaluate the model for 10 epochs. The results are shown in Fig.~\ref{fig2}.

From Fig.~\ref{fig2}, we can make several important observations.  1) Our approach can satisfy the requirements of different constraints (Fig.~\ref{fig2}, a) while completing the main task (Fig.~\ref{fig2}, b). 2) The tighter constraint given to the agent is, the lower is the total reward, as it influences the agent to imitate the demonstrated behavior, which will move it away from car 1 to get more auxiliary reward with the value of -0.01 (Fig.~\ref{fig2}, b). 3) The tighter constraint needs more time to adjust the value of \(\beta\) (Fig.~\ref{fig2}, c). 

These results show that our approach is efficient because it can automatically trade off task completion and behavioral imitation to find a policy that satisfies both.  

\section{Conclusion}
In this paper, we present Regularized Soft Actor-Critic (RSAC), which is formulated under Constrained Markov Decision process (CMDP) by combining the maximum entropy objective and the behavior imitation requirement. This algorithm allows the agent to complete the task with a specific behavior style. Our theoretical results derive regularized soft policy iteration, which we show to converge to optimal policy. By using approximate inference, we formulate a practical Regularized Soft Actor-Critic algorithm.  We evaluate our algorithm by having the agent perform the same task while imitating the demonstrated behavior to different degrees. The experimental results show that our method is effective.

In the future, we plan to apply our method to RPG video games. We want the agent to complete a task in a real environment, while at the same time mimicking the expert policy under ideal conditions to some degrees. Imitation of the expert strategy helps to accelerate the convergence of the model, and making the learned strategy different from the expert strategy in some degrees helps to find the optimal solution in the real environment.

\bibliography{main.bib}

\appendix
\section{Regularized Soft-Q learning}
\subsection{Regularized Soft-Q function}
We define the regularized soft state-action function \(Q_{rs}^{\pi}(s_t,a_t)\) as:
\begin{scriptsize}
\begin{equation} \label{eq23}
\begin{aligned}
& Q^*_{rs}(s_t,a_t) = r_t + E_{(s_{t+1},...)\sim\rho_\pi}\left[\sum\limits_{l=1}^{\infty}\gamma^l\left(r_{t+l}+\alpha H\left(\pi(\cdot|s_{t+l})\right) -\beta_{t+l} CE\left(\pi(\cdot|s_{t+l}), \pi^{d}(\cdot|s_{t+l})\right) \right)\right]\\
& =E_{s_{t+1}}\left[r_t+\gamma\left(\alpha H(\pi(\cdot|s_{t+1}))-\beta_{t+1}CE\left(\pi(\cdot|s_{t+1}), \pi^{d}(\cdot|s_{t+1})\right)+Q_{rs}^{\pi}(s_{t+1},a_{t+1})\right)\right]\\
\end{aligned}
\end{equation}
\end{scriptsize}
Then we can easily rewrite the objective in Eq.\ref{eq5} as
\begin{scriptsize}
\begin{equation} \label{eq24}
J(\pi)=\sum\limits_{t=0}^{T}E_{(s_t,a_t)\sim\rho_{\pi}}\left[Q_{rs}^{\pi}(s_t,a_t)+\alpha H\left(\pi(\cdot|s_t)\right)-\beta_tCE\left(\pi(\cdot|s_t),\pi^{d}(\cdot|s_t)\right)\right]
\end{equation}
\end{scriptsize}

\subsection{Policy improvement}
Given a policy \(\pi\) and the policy of demonstrated behavior \(\pi^{d}\), define a new policy \(\widetilde{\pi}\) as
\begin{scriptsize}
\begin{equation} \label{eq25}
\widetilde{\pi}\propto\left(\left(\pi^{d}(a_t|s_t)\right)^{\frac{\beta_t}{\alpha}}\exp\left(\frac{1}{\alpha}Q_{rs}^{\pi}(s_t,a_t)\right)\right)
\end{equation}
\end{scriptsize}
Assume that throughout our computation, Q is bounded and \((\pi^{d}(a_t|s_t))^{\frac{\beta_t}{\alpha}}\exp(\frac{1}{\alpha}Q_{rs}^{\pi}(s_t,a_t))\) is bounded for any \(s\) and \(a\), then \(Q_{rs}^{\widetilde{\pi}} \ge Q_{rs}^{\pi} \forall s,a.\)
\par The proof is based on the observation that
\begin{scriptsize}
\begin{equation} \label{eq26}
\begin{aligned}
& E_{a\sim\widetilde{\pi}}\left[Q_{rs}^{\pi}(s_t,a_t)\right]+\alpha H\left(\widetilde{\pi}(\cdot|s_t)\right)-\beta CE\left(\widetilde{\pi}(\cdot|s_t), \pi^{d}(\cdot|s_t)\right)\\
& \ge E_{a\sim\pi}\left[Q_{rs}^{\pi}(s_t,a_t)\right]+\alpha H\left(\pi(\cdot|s_t)\right)-\beta CE\left(\widetilde{\pi}(\cdot|s_t), \pi^{d}(\cdot|s_t)\right)
\end{aligned}
\end{equation}
\end{scriptsize}
\par The proof is straight-forward by noticing that
\begin{scriptsize}
\begin{equation} \label{eq27}
\begin{aligned}
& E_{a\sim\pi}\left[Q_{rs}^{\pi}(s_t,a_t)\right]+\alpha H\left(\pi(\cdot|s_t)\right)-\beta CE\left(\widetilde{\pi}(\cdot|s_t), \pi^{d}(\cdot|s_t)\right)\\
& =-\alpha D_{KL}\left(\pi(\cdot|s_t), \widetilde{\pi}(\cdot|s_t)\right)+\alpha log\sum\limits_{a'_t}\left(\pi^{d}(a'_t|s_t)^{\frac{\beta_t}{\alpha}}\exp{\left(\frac{1}{\alpha} Q^\pi_{rs}(s_t, a'_t)\right)} \right)
\end{aligned}
\end{equation}
\end{scriptsize}
Therefore, the LHS is only maximized if the KL Divergence on the RHS is minimized only when \( \widetilde{\pi}=\pi\).
\par Then we can continue show that
\begin{scriptsize}
\begin{equation} \label{eq28}
\resizebox{.9 \textwidth}{!}{%
$
\begin{aligned}
& Q_{rs}^{\pi}(s_t,a_t)=E_{s_1}\left[r_0+\gamma\left(\alpha H\left(\pi(\cdot|s_1)\right)-\beta_1 CE\left(\pi(\cdot|s_1), \pi^{d}(\cdot|s_1)\right)+E_{a_1\sim\pi}Q_{rs}^{\pi}(s_{1},a_{1})\right)\right]\\
& \leq E_{s_1}\left[r_0+\gamma\left(\alpha H\left(\widetilde{\pi}(\cdot|s_1)\right)-\beta_1 CE\left(\widetilde{\pi}(\cdot|s_1), \pi^{d}(\cdot|s_1)\right)+E_{a_1\sim\widetilde{\pi}}Q_{rs}^{\pi}(s_{1},a_{1})\right)\right]\\
& =E_{s_1}\left[r_0+\gamma\left(\alpha H\left(\widetilde{\pi}(\cdot|s_1)\right)-\beta_1 CE\left(\widetilde{\pi}(\cdot|s_1), \pi^{d}(\cdot|s_1)\right) + r_1\right)+\gamma^2\left(\alpha H\left(\pi(\cdot|s_2)\right)-\beta_2 CE\left(\pi(\cdot|s_2), \pi^{d}(\cdot|s_2)\right)+E_{a_2\sim\pi}Q_{rs}^{\pi}(s_{2},a_{2})\right)\right]\\
& \leq E_{s_1}\left[r_0+\gamma\left(\alpha H\left(\widetilde{\pi}(\cdot|s_1)\right)-\beta_1 CE\left(\widetilde{\pi}(\cdot|s_1), \pi^{d}(\cdot|s_1)\right) + r_1\right)+\gamma^2\left(\alpha H\left(\widetilde{\pi}(\cdot|s_2)\right)-\beta_2 CE\left(\widetilde{\pi}(\cdot|s_2), \pi^{d}(\cdot|s_2)\right)+E_{a_2\sim\widetilde{\pi}}Q_{rs}^{\pi}(s_{2},a_{2})\right)\right]\\
& \vdots\\
& \leq E_{\tau\sim\widetilde{\pi}}\left[r_0+\sum\limits_{t=1}^{\infty}\gamma^t\left(\alpha H\left(\widetilde{\pi}(\cdot|s_t)\right)-\beta_t CE\left(\widetilde{\pi}(\cdot|s_t),\pi^{d}(\cdot|s_t)\right)+r_t\right)\right]\\
& =Q_{rs}^{\widetilde{\pi}}\\
\end{aligned}%
$
}
\end{equation}
\end{scriptsize}
We can see that if we start from an arbitrary policy \(\pi_0\) and we define the policy iteration as 
\begin{scriptsize}
\begin{equation} \label{eq29}
\pi_{i+1}(\cdot|s)\propto\left(\left(\pi^{d}(\cdot|s)\right)^{\frac{\beta_i}{\alpha}}\exp\left(\frac{1}{\alpha}Q_{rs}^{\pi_i}(s,\cdot)\right)\right),\\
\end{equation}
\end{scriptsize}
then \(Q_{rs}^{\pi_i}(s,a)\) improve monotonically. Similar to \citep{Haarnoja_Tang}, with certain regularity conditions satisfied, any non-optimal policy can be improved this way.

\subsection{Regularized Soft Bellman Equation and Regularized Soft Value Iteration}
Recall the definition of regularized soft value function
\begin{scriptsize}
\begin{equation} \label{eq30}
V_{rs}^{\pi}(s_t)=\alpha log\sum\limits_{a'_t}\pi^{d}(a'_t|s_t)^{\frac{\beta_t}{\alpha}}\exp{\left(\frac{1}{\alpha} Q^\pi_{rs}(s_t, a'_t)\right)}.\\
\end{equation}
\end{scriptsize}
Suppose
\begin{scriptsize}
\begin{equation} \label{eq31}
\pi(a_t|s_t)=\left(\pi^{d}(a_t|s_t)\right)^{\frac{\beta_t}{\alpha}}exp{\left(\frac{1}{\alpha}Q^\pi_{rs}(s_t, a_t)-\frac{1}{\alpha}V_{rs}^{\pi}(s_t)\right)},\\
\end{equation}
\end{scriptsize}
then we can show that 
\begin{scriptsize}
\begin{equation} \label{eq32}
\begin{aligned}
& Q_{rs}^{\pi}(s,a)=r(s,a)+\gamma E_{s'\sim p_s}\left[\alpha H\left(\pi(\cdot|s')\right)-\beta CE\left(\pi(\cdot|s'),\pi^{d}(\cdot|s')\right)+E_{a'\sim\pi(\cdot|s')}\left[Q_{rs}^{\pi}(s',a')\right]\right]\\
& =r(s,a)+\gamma E_{s'\sim p_s}\left[V_{rs}^{\pi}(s')\right]\\
\end{aligned}
\end{equation}
\end{scriptsize}

We define the regularized soft value iteration operator \(\Gamma\) as
\begin{scriptsize}
\begin{equation} \label{eq33}
\Gamma Q(s,a)=r(s,a)+\gamma E_{s'\sim p_s}\left[\alpha log\sum\limits_{a'_t}\pi^{d}(a'_t|s_t)^{\frac{\beta_t}{\alpha}}\exp{\left(\frac{1}{\alpha} Q^\pi_{rs}(s_t, a'_t)\right)}\right].\\
\end{equation}
\end{scriptsize}
We can show that the operator defined above is a contraction mapping. We define a norm on Q-values \(||Q_1-Q_2||\triangleq Max_{s,a}|Q_1(s,a)-Q_2(s,a)|\). Suppose \(\varepsilon=||Q_1-Q_2||\), then
\begin{scriptsize}
\begin{equation} \label{eq34}
\begin{aligned}
& log\sum\limits_{a'_t}\pi^{d}(a'_t|s_t)^{\frac{\beta_t}{\alpha}}\exp{\left(\frac{1}{\alpha} {Q_1}_{rs}(s_t, a'_t)\right)}\leq log\sum\limits_{a'_t}\pi^{d}(a'_t|s_t)^{\frac{\beta_t}{\alpha}}\exp{\left(\frac{1}{\alpha} {Q_2}_{rs}(s_t, a'_t)+\varepsilon \right)}\\
& =log\sum\limits_{a'_t}\pi^{d}(a'_t|s_t)^{\frac{\beta_t}{\alpha}}\exp{\left(\frac{1}{\alpha}{Q_2}_{rs}(s_t, a'_t)\right)} * \exp{(\varepsilon)}\\
& =\varepsilon+log\sum\limits_{a'_t}\pi^{d}(a'_t|s_t)^{\frac{\beta_t}{\alpha}}\exp{\left(\frac{1}{\alpha} {Q_2}_{rs}(s_t, a'_t)\right)}\\
\end{aligned}
\end{equation}
\end{scriptsize}
Similarly, 
\begin{scriptsize}
\(log\sum\limits_{a'_t}\pi^{d}(a'_t|s_t)^{\frac{\beta_t}{\alpha}}\exp{\left(\frac{1}{\alpha} {Q_1}_{rs}(s_t, a'_t)\right)}\ge -\varepsilon+log\sum\limits_{a'_t}\pi^{d}(a'_t|s_t)^{\frac{\beta_t}{\alpha}}\exp{\left(\frac{1}{\alpha} {Q_2}_{rs}(s_t, a'_t)\right)}\).\\
\end{scriptsize}
Therefore \begin{scriptsize}\(||\Gamma Q_1-\Gamma Q_2||\leq \gamma\varepsilon=\gamma||Q-1-Q_2||\)\end{scriptsize}. So \(\Gamma\) is a contraction.

\end{document}